\pdfoutput=1
\documentclass[11pt]{article}

\usepackage[preprint]{acl}

\usepackage{times}
\usepackage{latexsym}

\usepackage[T1]{fontenc}

\usepackage[utf8]{inputenc}

\usepackage{microtype}

\usepackage{inconsolata}

\usepackage{graphicx}
\usepackage{url}
\usepackage{booktabs}
\usepackage{subcaption}
\usepackage{adjustbox}

\usepackage{xcolor}
\definecolor{nicegreen}{RGB}{192,239,171}
\definecolor{niceblue}{RGB}{174,224,246}
\definecolor{nicered}{RGB}{239,171,192}

\title{\textsc{ReTraceQA}: Evaluating Reasoning Traces of Small Language Models in Commonsense Question Answering}

\author{Francesco Maria Molfese\thanks{\ Equal contribution.}, Luca Moroni\footnotemark[1], Ciro Porcaro, Simone Conia, Roberto Navigli \\
	Sapienza NLP Group, Sapienza University of Rome \\
	\texttt{\{molfese, moroni, porcaro, conia, navigli\}@diag.uniroma1.it}
}

\begin{document}
\maketitle

\begin{abstract}
While Small Language Models (SLMs) have demonstrated promising performance on an increasingly wide array of commonsense reasoning benchmarks, current evaluation practices rely almost exclusively on the accuracy of their final answers, neglecting the validity of the reasoning processes that lead to those answers.
To address this issue, we present \textsc{ReTraceQA}, a novel benchmark that introduces process-level evaluation for commonsense reasoning tasks.
Our expert-annotated dataset reveals that in a substantial portion of instances (14-24\%), SLMs provide correct final answers despite flawed reasoning processes, suggesting that the capabilities of SLMs are often overestimated by evaluation metrics that focus only on comparing the final answer with the ground truth.
Indeed, we show that, when employing strong Large Language Models (LLMs) as automated judges for reasoning-aware evaluation rather than answer-only metrics, SLM performance drops significantly across all models and datasets, with scores decreasing by up to 25\%.
\end{abstract}

\section{Introduction}
Recent work in language modeling has led to effective SLMs with impressive performance levels across various benchmarks~\cite{grattafiori2024llama3herdmodels, abdin2024phi4technicalreport, open-llm-leaderboard-v2, qwen2025qwen25technicalreport}.
However, current evaluation practices rely almost exclusively on final answer accuracy, i.e., counting an instance as correct when the model's prediction matches the ground truth, regardless of the reasoning process.
This answer-centric approach overlooks a fundamental factor: models can arrive at correct answers through invalid reasoning paths, artificially inflating performance metrics and masking important weaknesses in their actual reasoning capabilities.

To this end, the research community has recently proposed several benchmarks to examine ``reasoning traces'' -- the step-by-step explanations generated by language models to arrive at their final answers -- in a more systematic way~\cite{zheng2024processbenchidentifyingprocesserrors, zeng2024mrgsm8kmetareasoningbenchmarklarge, zeng2024mrbenmetareasoningbenchmarkevaluating, tyen-etal-2024-llms}.
These benchmarks are necessary for the development and evaluation of automatic approaches, such as Process Reward Models (PRMs)~\cite{lightman2023letsverifystepstep,wang2024mathshepherdverifyreinforcellms, zhang2025lessonsdevelopingprocessreward} and LLMs as judges \cite{gu2025surveyllmasajudge}, aimed at identifying the specific location of errors within reasoning traces and not just the correctness of the final answer~\cite{zheng2024processbenchidentifyingprocesserrors, zeng2024mrgsm8kmetareasoningbenchmarklarge, zeng2024mrbenmetareasoningbenchmarkevaluating, tyen-etal-2024-llms}.

However, contemporary work faces two key limitations. First, existing benchmarks focus primarily on mathematics and science, leaving reasoning processes in other areas like commonsense reasoning largely underexplored, despite requiring fundamentally different capabilities. 
Second, specialized PRMs and LLMs employed as judges are typically used to optimize task performance through feedback during fine-tuning or Best-of-N sampling, rather than evaluating whether reasoning traces that reach correct answers contain intermediate errors, potentially leading to inflated performance assessments.
Therefore, our research question is: \textit{how can we effectively evaluate reasoning processes in commonsense, and to what extent do current answer-only metrics misrepresent SLM capabilities?}

To address these limitations, we provide the following contributions: 
\begin{itemize}
    \item We introduce \textsc{ReTraceQA}, the first benchmark for evaluating reasoning traces of SLMs in commonsense reasoning tasks, including a set of 2,421 reasoning traces manually annotated with step-level error locations and qualitative error categorizations;
    \item Quantitative evidence that up to 24\% of flawed reasoning traces still produce the correct final answer, demonstrating how current answer-only evaluations significantly overestimate model capabilities;
    \item Comprehensive reference-based evaluation of both closed and open-source LLMs as judges, revealing that, while models can often detect whether a trace is correct as a whole, they struggle to identify the exact location of reasoning errors;
    \item Reference-free evaluation of LLM-as-a-judge models and mathematical PRMs applied to commonsense reasoning, revealing substantial performance degradation when transferring across domains.
\end{itemize}
Our findings show that answer-only metrics substantially overestimate SLM performance, with scores dropping by up to 25\% when accounting for reasoning correctness, also highlighting the need for reasoning-aware evaluation beyond STEM domains.
\textsc{ReTraceQA} provides both a practical benchmark and strong evidence that current evaluation practices can misrepresent reasoning in commonsense question answering tasks. 
We release our benchmark, code, and data at \url{https://github.com/SapienzaNLP/ReTraceQA}.

\section{Related Work}

\paragraph{Process-Based Evaluation Approaches.}
The research community has introduced two main approaches for assessing reasoning quality beyond final answers: Process Reward Models (PRMs) and LLM-as-a-judge.
PRMs are specialized models fine-tuned to evaluate the correctness of reasoning steps, in contrast to Outcome Reward Models (ORMs), which focus solely on final answers~\cite{uesato2022solvingmathwordproblems, lightman2023letsverifystepstep}. PRMs specifically aim to identify the first erroneous step in a reasoning trace, enabling both targeted feedback for model training and quality filtering in Best-of-N selection scenarios~\cite{pan2023letsreinforcestepstep, wang2024mathshepherdverifyreinforcellms}.
PRMs can be built and trained in several ways: \citet{lightman2023letsverifystepstep} used human-labeled data for error detection, while \citet{li-etal-2023-making} and \citet{wang2024mathshepherdverifyreinforcellms} employed Monte Carlo estimation to determine the probability of a chain of steps to be correct.
More recent work by \citet{hosseini2024vstartrainingverifiersselftaught} and \citet{zhang2025lessonsdevelopingprocessreward} leverages larger LLMs as automated judges to generate training signals for PRMs, creating a teacher-student paradigm for reasoning evaluation.

In parallel to specialized PRMs, general-purpose LLMs prompted as judges have emerged as an effective –– albeit expensive –– alternative approach.
These models assess reasoning trace validity without task-specific training, providing both binary correctness judgments and localized error identification~\cite{zheng2024processbenchidentifyingprocesserrors}.
While more flexible than PRMs, judges may lack the specialization that targeted training provides.

\paragraph{Benchmarks for Reasoning Evaluation.}
Several benchmarks have been developed to evaluate models' abilities to identify errors in reasoning traces, each with distinct characteristics.
ProcessBench~\cite{zheng2024processbenchidentifyingprocesserrors} specifically targets reasoning error identification by requiring models to indicate the exact location of incorrect steps within mathematical reasoning traces.
MR-Ben and MR-GSM8K~\cite{zeng2024mrgsm8kmetareasoningbenchmarklarge, zeng2024mrbenmetareasoningbenchmarkevaluating} offer more comprehensive meta-reasoning assessment, including error localization, error explanation, and suggested corrections.
Findings from these benchmarks consistently demonstrate that even state-of-the-art LLMs struggle to detect reasoning error locations accurately, though they do show potential for providing helpful corrections once errors have been explicitly identified~\cite{tyen-etal-2024-llms, huang2024largelanguagemodelsselfcorrect}.

\paragraph{Limitations and Research Gaps.}
Despite progress in process-based evaluation, existing work presents three key limitations. 
First, PRMs and judge models tend to be used primarily for Best-of-N selection, thus for ranking multiple outputs in order to improve generation rather than as tools for validating reasoning traces during evaluation. 
Second, current benchmarks are largely restricted to mathematical domains, overlooking reasoning types found in commonsense tasks that involve qualitatively different inference. 
Third, the implications of reasoning-aware evaluation on SLM assessment remain underexplored, particularly in respect of how final answer metrics can misrepresent underlying reasoning quality.

\begin{figure*}[t]
    \centering
    \includegraphics[width=\textwidth]{pipeline_fig.png}
    \caption{Overview of the \textsc{ReTraceQA} pipeline (top) and a representative example of a fully-annotated instance (bottom). The bottom panel illustrates the identification of error location and categorization, tracing the transformation from the initial raw model output to the final reformatted and structured reasoning trace.}
    \label{fig:retraceqa}
\end{figure*}

\section{ReTraceQA}\label{sec:benchmark}
In this section, we introduce \textsc{ReTraceQA}, our novel gold benchmark designed to assess the ability of LLMs to determine whether a reasoning trace of an SLM is correct, or if it is not, to identify the specific step where an error occurs.
In the following, Section \ref{sec:task} provides the formal task definition, Section \ref{sec:dataset-selection} describes the datasets selected for the benchmark, Section \ref{sec:solution-generation} explains how reasoning traces are generated using a range of SLMs, Section \ref{sec:solution-reformatting} explains how we divide the reasoning traces into discrete steps, Section \ref{sec:human-annotation} details the human annotation process used to construct the resulting resource, and, finally, Section \ref{sec:benchmark-stats} presents descriptive statistics of our benchmark.
Figure \ref{fig:retraceqa} shows the overview of the whole pipeline together with an example of annotation.

\subsection{Task Definition}\label{sec:task}
Given a commonsense reasoning question, the goal is to evaluate the validity of an SLM-generated reasoning trace by identifying the earliest step at which an error occurs, if any.
Formally, let $q$ denote the input question (with an optional set of choices $C$), and let $S = [s_0, s_1, \ldots, s_n]$ represent the step-by-step reasoning trace. 
The task is to predict an index $i \in \{-1, 0, \ldots, n\}$, where $i = -1$ signifies that all reasoning steps are correct, and $i \ge 0$ indicates that the first error occurs at step $s_i$. 
This formulation is in line with state-of-the-art benchmarks like ProcessBench \cite{zheng2024processbenchidentifyingprocesserrors}, in which the authors note that, for steps after the first error, the meaning of their correctness may become ambiguous or
debatable.
Indeed, derivations based on incorrect premises can make sense, but still remain on a globally incorrect
reasoning path \cite{lightman2023letsverifystepstep}.
Based on this assumption, we choose to focus on
identifying the earliest-occurring error in the reasoning traces.

\subsection{Dataset Selection}\label{sec:dataset-selection}
To construct our benchmark, we source questions from four widely-used datasets in commonsense reasoning: CommonsenseQA \cite[CSQA]{talmor-etal-2019-commonsenseqa}, OpenBookQA \cite[OBQA]{mihaylov-etal-2018-suit}, QASC \cite{khot2020qascdatasetquestionanswering}, and StrategyQA \cite{geva-etal-2021-aristotle}, all of which are multiple-choice or binary question answering datasets that provide a question along with a set of candidate answers.
These datasets primarily target commonsense reasoning grounded in general world knowledge, but also feature questions involving encyclopedic and subject-specific knowledge, as well as reasoning over spatial, temporal, or causal relationships. 
When test set labels are not publicly available, we follow standard practice and instead utilize the corresponding development sets \cite{liu-etal-2023-crystal, molfese-etal-2024-zebra}; this applies to CSQA and QASC. Together, these datasets span a range of reasoning challenges across commonsense domains, making them well-suited for evaluating the correctness and robustness of reasoning traces.

\subsection{Solution Generation}\label{sec:solution-generation}
For each instance in our selected datasets, we generate step-by-step reasoning traces using SLMs from the widely used LLaMA, Phi and Qwen families of open-source language models \cite{grattafiori2024llama3herdmodels, abdin2024phi4technicalreport, qwen2025qwen25technicalreport}.
We follow standard practice and define an SLM as any language model with no more than 10 billion parameters \cite{fu2023specializingsmallerlanguagemodels, wang2024comprehensivesurveysmalllanguage}.
Specifically, we use the following instruction-tuned variants: Qwen2.5-1.5B-Instruct, Qwen2.5-3B-Instruct, Qwen2.5-7B-Instruct, Llama-3.2-1B-Instruct, Llama-3.2-3B-Instruct, Llama-3.1-8B-Instruct, and Phi-4-mini-instruct. 
This selection enables us to examine performance variation within model families as model size increases (with the exception of Phi, which is only available in a single size under 10 billion parameters), while also capturing differences across architectures. 
We generate traces by prompting models with a zero-shot Chain-of-Thought (CoT) setup \cite{wei2023chainofthoughtpromptingelicitsreasoning, kojima2023largelanguagemodelszeroshot}, which encourages step-by-step reasoning without conditioning on specific examples.

Initially, we collect a total of 3,334 original questions distributed across the datasets as follows: 1,221 questions from CSQA, 500 questions from OBQA, 926 questions from QASC and 687 questions from StrategyQA.
For each original question, we generate reasoning traces using 7 distinct SLMs, resulting in an initial pool of 23,338 total reasoning traces.
Then, we perform careful sampling from this initial pool to ensure three factors simultaneously: (i) balanced representation of correct and incorrect traces in terms of final answer accuracy, (ii) balanced representation of each model and (iii), uniqueness of each question.
This sampling step reduces the dataset to a total of 2,779 unique instances (i.e., each instance is a unique question associated with exactly one reasoning trace).
Details on reasoning trace generation and answer-based classification can be found in Appendices \ref{app:cot-prompts} and \ref{app:answer-extractor}, respectively.

\subsection{Solution Reformatting}\label{sec:solution-reformatting}
A key step in building \textsc{ReTraceQA} involves ensuring that model-generated reasoning traces are segmented into coherent, interpretable steps. In mathematical domains, prior work has shown that automatic solutions often require post-hoc re-segmentation due to formatting inconsistencies and unclear boundaries between reasoning steps \cite{zheng2024processbenchidentifyingprocesserrors}. 
In contrast, we find that, for commonsense reasoning tasks, step segmentation emerges more naturally.
When using CoT prompting, models tend to produce clearly delineated reasoning traces, with each step expressed as a self-contained sentence, making reformatting unnecessary.\footnote{We split reasoning traces at ``\textbackslash n\textbackslash n'', falling back to ``\textbackslash n'' only when the double-newline sequence is not found.}

\begin{table*}[t]
    \begin{adjustbox}{width=\textwidth}
        \footnotesize
        \centering
        \begin{tabular*}{\textwidth}{p{0.25\textwidth}|p{0.50\textwidth}|p{0.17\textwidth}}
            \toprule
            \textbf{Input} & \textbf{Reasoning Trace} & \textbf{Error Category} \\ 
            \midrule
            \parbox[t]{0.25\textwidth}{
                \textbf{Question}: Hand dryers can also be used to \\
                \textbf{Choices}: \\
                A. keep cold drinks cool \\
                \underline{B. dry out wet clothes} \\
                C. hydrate your face and hands \\
                D. make a damp rag damper \\
            } &
            \parbox[t]{0.50\textwidth}{                
                \textbf{Paragraph 0:} \par
                To answer this question, I'll analyze each option and eliminate the incorrect ones based on the properties and common uses of hand dryers. \par
                [...] \par
                \textbf{Paragraph 5:} \par
                \parbox{0.49\textwidth}{
                    \textcolor{red}{I will choose option E, which is not provided in the list. Since the correct answer is not among the options, I will choose the one that is least likely to be correct.}
                }
            } &
            \parbox[t]{0.17\textwidth}{Misinterpretation.} \\

            \midrule
            
            \parbox[t]{0.25\textwidth}{
                \textbf{Question}: Where can bees find the materials needed to make honey? \\
                \textbf{Choices}: \\
                A. heated by the sun \\
                B. occurs over a wide range \\
                \underline{C. trees and flowers} \\
                D. buildings \\
                E. hives \\
                F. leaves \\
                G. 70-75 degrees Fahrenheit \\
                H. interior chambers
            } &
            \parbox[t]{0.50\textwidth}{             
                \textbf{Paragraph 0:} \par
                To answer this question, I need to consider the characteristics of honey and its production process. Honey is produced by honeybees (Apis mellifera) through a process called nectar collection and regurgitation \par
                [...] \par
                \textbf{Paragraph 5:} \par
                \parbox{0.49\textwidth}{
                    \textcolor{red}{F. leaves: Leaves are the primary source of nectar for honey production.}
                }
            } & 
            \parbox[t]{0.17\textwidth}{Hallucination.} \\

            \midrule

            \parbox[t]{0.25\textwidth}{
                \textbf{Question}: During the winter hunt he could hear every motion in the woods, this was because of the what of everything? \\
                \textbf{Choices}: \\
                A. being still \\
                B. silence \\
                C. stationary \\
                \underline{D. stillness} \\
                E. standing still
            } &
            \parbox[t]{0.50\textwidth}{
                \textbf{Paragraph 0:} \par
                The question states that the person can hear every motion in the woods, which implies that there is something that allows them to be aware of even the slightest movements. \par
                [...] \par
                \textbf{Paragraph 3:} \par
                \parbox{0.49\textwidth}{
                    \textcolor{red}{The word "being still" (A) implies that the person is not moving, which would make it harder to hear movements, not easier.}
                }
            } & 
            \parbox[t]{0.17\textwidth}{Reasoning.} \\
            \bottomrule
        \end{tabular*}
    \end{adjustbox}
    \caption{Examples of the three error categories identified in \textsc{ReTraceQA}. For each instance, we present the input question and choices, the segmented reasoning trace generated by the SLM, and the final error categorization.}
    \label{tab:reasoning-trace-errors}
\end{table*}

\subsection{Human Annotation}\label{sec:human-annotation}

To construct a benchmark that enables both binary reasoning evaluation and fine-grained error localization, we annotate a diverse set of SLM-generated reasoning traces with step-level error information.
The annotation task follows the setup described in Section \ref{sec:task}: for each reasoning trace, annotators are asked to identify the earliest step that contains an error, or to indicate that the entire trace is correct. 
Additionally, annotators are asked to assign one of three available labels classifying the nature of the error.
To ensure high-quality reproducible labels and to minimize the inherent subjectivity in evaluating commonsense, we employ a hierarchical error taxonomy.
We categorize errors from grounding to inference, ensuring that each error is assigned to a single, mutually exclusive category based on its primary cause.
Specifically, a step is considered erroneous if it falls into one of the following categories: 

\begin{itemize}
    \item \textbf{Misinterpretation (Grounding Level):} errors occurring when the model misunderstands the question, choice meanings, or task requirements. This includes misrepresenting previous steps, referencing non-existent choices, or providing multiple answers.
    \item \textbf{Hallucination (Content Level):} errors involving the introduction of empirically false or unverifiable world knowledge. This category is strictly reserved for instances where the model's logic might be structurally sound, but the ``factual'' building blocks are incorrect (e.g., stating ``wolves are not found in arctic regions''). A step is labeled as a hallucination only if the input was correctly interpreted, but the world knowledge used by the model is false or unverifiable.
    \item \textbf{Reasoning (Inference Level):} errors involving the logical transitions between or within steps. This occurs when the model connects correct premises or facts using an invalid logical leap. For example, a model might correctly state that ``salt lowers the freezing point of water'' but incorrectly infer that ``this makes it easier for ice to form.'' This label is applied when the factual content of the step is correct, but the deduction is logically unsound.
\end{itemize}

Three expert annotators with PhD-level backgrounds in computer science or linguistics -- who are also the authors of this paper -- perform the annotation. Each annotator is given the SLM-generated reasoning trace, the original question, optional answer choices and supporting facts, and the gold answer from the dataset, and is instructed to judge correctness based solely on the reasoning trace, not on the final answer alone.
Annotators are also asked to flag problematic instances using a dedicated \textsc{Invalid} tag. These include: ambiguous questions with multiple plausible answers, grammatical or structural issues that impair interpretation and labeling inconsistencies in the original dataset (e.g., an incorrect gold answer).
To safeguard the quality of the final benchmark, we exclude all flagged instances. This results in a total of 2,421 clean and fully annotated examples in \textsc{ReTraceQA}.

To evaluate annotation consistency, we randomly sample 25 instances from each of the four datasets in the benchmark (100 total). 
All three annotators independently label this subset following the same guidelines. Inter-annotator agreement, measured via Fleiss’s kappa, yields a score of 0.84, indicating an \textit{``almost perfect''} agreement according to standard interpretation~\cite{Landis1977}.

Figure \ref{fig:retraceqa} illustrates the transition from raw model output to a fully annotated instance, depicting the reformatted solution alongside its error location and categorization. 
To further clarify our taxonomy, Table \ref{tab:reasoning-trace-errors} provides representative examples for each of the three error categories, with an additional set of cases detailed in Appendix \ref{app:error-examples}. 
A comprehensive and detailed description of the annotation protocol is available in Appendix \ref{app:annotation-guidelines}.

\subsection{Benchmark Statistics}\label{sec:benchmark-stats}
Table \ref{tab:reasoning-trace-analysis} presents a detailed analysis of the reasoning traces across each subset, including the number of samples that reach a correct or incorrect final answer, the proportion of process errors (instances with correct answers but flawed reasoning), descriptive statistics on reasoning trace length and the percentage of instances falling in each of the available error categories.

A key observation is that a non-trivial percentage of responses, averaging to 17.9\% across datasets, arrive at the correct final answer despite containing a reasoning error. 
This pattern is consistent with findings from mathematical reasoning benchmarks \cite{zheng2024processbenchidentifyingprocesserrors}, in which even strong language models are able to reach the correct answer while making mathematical mistakes, and highlights a critical limitation of standard evaluation practices, which often overlook flawed intermediate reasoning when only final answers are assessed. 
As a result of such oversight, leaderboard metrics may overestimate the true reasoning capability of language models.
Moreover, we can see a consistent distribution of error categories across the four subsets of our benchmark.
Specifically, hallucination errors constitute the majority of failures (41.9\%--62.5\%), followed by reasoning errors (27.9\%--35.4\%) and misinterpretation errors (9.6\%--24.1\%).
In Figure \ref{fig:avg-error-type} we report the error category distribution averaged across the four subsets of our benchmark.
This suggests that SLMs struggle primarily with factual grounding, frequently generating unverifiable claims or incorrect assumptions, though logical coherence issues also remain prevalent, accounting for roughly one-third of all errors.
The lower proportion of misinterpretation errors indicates that models generally understand task requirements and question semantics, but fail both in anchoring their reasoning in accurate world knowledge and in maintaining sound logical inference chains.
Individual statistics for each model are provided in Appendix \ref{app:model-stats}.

\begin{table}[t]
    \centering
    \small
    \resizebox{\columnwidth}{!}{
    \begin{tabular}{lcccc}
    \toprule
    & \textbf{CSQA} & \textbf{OBQA} & \textbf{QASC} & \textbf{StrategyQA} \\
    \midrule
    Final samples (error)      & 296 & 184 & 219 & 271 \\
    Final samples (correct)    & 603 & 244 & 245 & 359 \\
    \textbf{Total samples}     & \textbf{899} & \textbf{428} & \textbf{464} & \textbf{630} \\
    \midrule
    Process errors (\%)        & 16.3 & 14.7 & 16.6 & 24.0 \\
    Invalid instances          & 238 & 20 & 46 & 54 \\
    \midrule
    Avg. steps (error)        & 8.2 & 8.1 & 8.0 & 6.9 \\
    Avg. steps (correct)      & 8.2 & 7.8 & 7.9 & 6.8 \\
    \midrule
    \multicolumn{5}{c}{\textit{Error Category Distribution (\%)}} \\
    \cmidrule(lr){1-5}
    Hallucination             & 41.9 & 46.7 & 47.5 & 62.5 \\
    Reasoning                 & 34.0 & 34.7 & 35.4 & 27.9 \\
    Misinterpretation         & 24.1 & 18.6 & 17.1 & ~~9.6 \\
    \bottomrule
    \end{tabular}
    }
    \caption{\textsc{ReTraceQA} statistics. Process errors refer to instances with correct final answers but flawed reasoning. Invalid instances were flagged during annotation and excluded. Error category distributions are calculated over all erroneous traces.}
    \label{tab:reasoning-trace-analysis}
\end{table}

\begin{figure}[t]
    \centering
    \includegraphics[scale=0.2]{avg-error-type.png}
    \caption{Error category distribution averaged across the four subsets of our ReTraceQA benchmark.}
    \label{fig:avg-error-type}
\end{figure}

\section{Experimental Setup}
Our benchmark evaluates LLMs along two axes: (1) reference-free assessment of SLM reasoning trace validity in order to determine whether models can reliably provide fine-tuning feedback or perform Best-of-N selection without ground truth labels, and (2) reference-based assessment where models judge reasoning traces using both the correct answer and reasoning process, extending evaluation beyond final answer correctness alone. In the following, we list the models used for our experiments (Section \ref{sec:exp-models}) and the evaluation metrics for both reference-free and reference-based settings (Section \ref{sec:exp-metrics}).

\subsection{Models}\label{sec:exp-models}
\paragraph{LLM-as-a-judge.}
We follow recent work on automated evaluation~\cite{zheng2023judgingllmasajudgemtbenchchatbot} by prompting LLMs to assess SLM reasoning traces.
The prompt is slightly adapted from prior work \cite{zheng2024processbenchidentifyingprocesserrors} to better suit commonsense reasoning tasks (Appendix~\ref{app:judge-prompts}).
We evaluate the following set of open-weight and closed models: Mistral-Small-24B-Instruct-2501 \cite{mistral2025small3}, Llama-3.3-70B-Instruct \cite{grattafiori2024llama3herdmodels}, Qwen2.5-72B-Instruct \cite{qwen2025qwen25technicalreport}, Gemini-2.0-Flash \cite{google2025gemini2flash}, DeepSeek-R1 \cite{deepseekai2025deepseekr1incentivizingreasoningcapability}, GPT-4o-mini \cite{openai2024gpt4ocard}, GPT-4o \cite{openai2024gpt4ocard} and o1-mini \cite{openai2024openaio1card}. 
Greedy decoding is used for all models except o1-mini and DeepSeek-R1, for which we report performance using a sample at temperature 1.0 due to API constraints.

\paragraph{Process Reward Models.}
We evaluate several publicly available PRMs by extracting their step-wise correctness predictions and identifying the first step flagged as incorrect. 
The evaluated models fall into three groups:  
(1) math-shepherd-mistral-7B \cite{wang2024mathshepherdverifyreinforcellms}, which uses empirical correctness likelihoods over reasoning steps;  
(2) Skywork-o1-Open-PRM-Qwen-2.5-1.5B and Skywork-o1-Open-PRM-Qwen-2.5-7B \cite{skyworkopeno12024}, which output raw scalar scores;  
(3) Qwen2.5-Math-7B-PRM800K and Qwen2.5-PRM-7B \cite{zheng2024processbenchidentifyingprocesserrors}, fine-tuned respectively on the PRM800K dataset and on synthetic data derived from LLM-as-a-judge annotations.

For models in groups (1) and (3), trained with sigmoid activations over each step, we determine step correctness by rounding predictions to the nearest integer (1 = correct, 0 = incorrect). For models in group (2), we select a threshold that maximizes F1 on a validation split of CSQA, following \citet{zheng2024processbenchidentifyingprocesserrors}, and use it to round scalar scores.

\subsection{Evaluation Metrics}\label{sec:exp-metrics}
\paragraph{Reasoning Trace Evaluation.} 
For both reference-free and reference-based settings, we evaluate models using two complementary metrics: \textit{correct}, measuring accuracy in identifying fully valid traces (human-labeled as $-1$), and \textit{error}, measuring accuracy in localizing the first erroneous step in flawed traces (human-labeled as $i$, where $i \ge 0$). These metrics assess whether LLMs can provide targeted feedback to SLMs during training and quantify their reliability for Best-of-N scoring during evaluation. Following prior work~\cite{zheng2024processbenchidentifyingprocesserrors}, we report the harmonic mean (F1) of \textit{correct} and \textit{error} to balance overly permissive versus overly critical model behaviors.

\paragraph{Downstream SLM Evaluation.}
To assess the impact of reasoning-aware evaluation on SLMs using LLM-as-a-judge, we employ the best-performing judge from the reference-based evaluation on \textsc{ReTraceQA} under two configurations: (1) answer-only evaluation (simulating standard approaches) and (2) full trace validation (accepting predictions only when both reasoning and answers are correct). We measure performance using \textit{accuracy} (correctly distinguishing valid from invalid traces: $i = -1$ vs. $i \neq -1$) and \textit{error recall} (identifying flawed traces where both model and human annotations indicate $i \neq -1$).  We evaluate seven SLMs with the same judge under the two settings across four commonsense datasets, generating diverse reasoning traces at temperature 0.7 to ensure variety while avoiding overlap with our annotated benchmark.

\section{Results}
In this section, we first present LLM performance on reference-free and reference-based evaluation on \textsc{ReTraceQA} (Section \ref{sec:res-trace-eval}).
We then report downstream SLM evaluation results, where the best-performing LLM-as-a-judge on \textsc{ReTraceQA} is used to assess SLM-generated outputs across multiple benchmarks (Section \ref{sec:slm_eval}).

\begin{table*}[t]
\centering
\small
\resizebox{\textwidth}{!}{
\begin{tabular}{l|ccc|ccc|ccc|ccc|c}
\toprule
\textbf{Model} &
\multicolumn{3}{c|}{\textbf{CSQA}} &
\multicolumn{3}{c|}{\textbf{OBQA}} &
\multicolumn{3}{c|}{\textbf{QASC}} &
\multicolumn{3}{c|}{\textbf{StrategyQA}} &
\textbf{Avg. F1} \\
& correct & error & F1 & correct & error & F1 & correct & error & F1 & correct & error & F1 & \\
\midrule
\multicolumn{14}{c}{\textit{Process Reward Models (reference-free evaluation)}} \\
\midrule
Math-Shepherd-PRM-7B         & 95.3 & ~~4.2  & ~~8.0   & 92.4 & ~~6.1  & 11.5  & 70.2 & 10.2 & 17.9  & 58.7 & 18.7 & 28.4 & 16.5 \\
Skywork-PRM-1.5B             & \textbf{97.5} & ~~1.9  & ~~3.7   & \textbf{95.6} & ~~2.5  & ~~4.8   & \textbf{88.9} & ~~2.7  & ~~5.3   & 93.0 & ~~1.9  & ~~3.8  & ~~4.4 \\
Skywork-PRM-7B               & 83.0 & ~~9.1  & 16.4  & 77.6 & 11.0 & 19.3  & 57.3 & 10.6 & 17.9  & 86.4 & ~~6.2  & 11.6 & 16.3 \\
Qwen2.5-Math-7B-PRM800K      & 89.6 & 13.8 & 23.8  & 88.5 & 20.8 & 33.7  & 73.7 & 24.6 & 36.9  & \textbf{97.2} & ~~8.9  & 16.3 & 27.7 \\
Qwen2.5-Math-PRM-7B          & 86.8 & \textbf{20.9} & \textbf{33.8}  & 81.4 & \textbf{28.9} & \textbf{42.8}  & 70.2 & \textbf{37.2} & \textbf{48.6}  & 79.8 & \textbf{24.5} & \textbf{37.4} & \textbf{40.7} \\
\midrule
\textbf{Average}      & 90.4 & 10.0 & 17.1  & 87.1 & 13.9 & 22.4  & 72.1 & 17.1 & 25.3  & 83.0 & 12.0 & 19.5 & 21.1 \\
\midrule
\multicolumn{14}{c}{\textit{LLM-as-a-judge (reference-free evaluation)}} \\
\midrule
Mistral-Small-24B-Instruct   & 25.7 & 48.3 & 33.6  & 34.9 & 53.1 & 42.2  & 22.8 & 50.9 & 31.5  & 24.4 & 43.6 & 31.3 & 34.7 \\
LLaMA-3.3-70B-Instruct       & \textbf{87.2} & 33.3 & 48.2  & 87.9 & 44.5 & 59.1  & \textbf{83.6} & 38.9 & 53.1  & \textbf{91.1} & 32.6 & 48.0 & 52.1 \\
Qwen2.5-72B-Instruct         & 85.9 & 44.1 & 58.3  & 80.3 & 53.5 & 64.2  & 77.8 & 43.0 & 55.4  & 84.5 & 45.1 & \textbf{58.8} & 59.2 \\
DeepSeek-R1            & 56.6 & \textbf{57.1} & 56.8  & 49.7 & \textbf{62.5} & 55.4  & 52.6 & \textbf{55.6} & 54.1  & 44.1 & \textbf{63.5} & 52.1 & 54.6 \\
Gemini-2.0-Flash             & 82.9 & 46.2 & 59.3  & 88.5 & 57.9 & \textbf{70.1}  & 77.2 & 51.2 & 61.6  & 87.3 & 40.8 & 55.6 & 61.7 \\
GPT-4o                       & 86.4 & 47.8 & \textbf{61.5}  & \textbf{89.6} & 52.2 & 66.0  & 79.5 & 52.9 & \textbf{63.5}  & 89.7 & 39.1 & 54.4 & 61.4 \\
GPT-4o-mini                  & 62.5 & 47.1 & 53.7  & 71.0 & 54.3 & 61.5  & 49.1 & 49.5 & 49.3  & 75.1 & 39.8 & 52.0 & 54.1 \\
o1-mini                      & 82.3 & 46.9 & 59.7  & 79.8 & 61.6 & 69.5  & 73.7 & 54.9 & 62.9  & 77.0 & 45.1 & 56.9 & \textbf{62.3} \\
\midrule
\textbf{Average} & 71.2 & 46.3 & 53.9  & 72.8 & 54.9 & 61.0  & 64.6 & 49.6 & 53.9  & 71.7 & 43.7 & 51.1 & 55.0 \\
\midrule
\multicolumn{14}{c}{\textit{LLM-as-a-judge (reference-based evaluation)}} \\
\midrule
Mistral-Small-24B-Instruct   & 22.3 & 54.3 & 31.7  & 32.8 & 62.4 & 43.0  & 21.6 & 59.4 & 31.7  & 14.6 & 55.4 & 23.1 & 32.4 \\
LLaMA-3.3-70B-Instruct       & 87.7 & 45.2 & 59.7  & \textbf{90.7} & 52.2 & 66.3  & 85.7 & 54.3 & 66.5  & 82.6 & 56.6 & 67.2 & 64.9 \\
Qwen2.5-72B-Instruct         & \textbf{89.6} & 50.6 & 64.7  & 85.3 & 59.2 & 69.9  & \textbf{86.6} & 58.4 & 69.7  & 83.6 & 56.4 & 67.3 & 67.9 \\
DeepSeek-R1                  & 53.8 & \textbf{61.5} & 57.4  & 49.2 & 66.1 & 56.4  & 49.7 & 65.9 & 56.7  & 37.6 & 63.6 & 47.2 & 54.4 \\
Gemini-2.0-Flash             & 86.6 & 52.2 & 65.2  & 89.1 & 64.1 & 74.5  & 78.9 & 60.4 & 68.4  & 66.2 & 59.0 & 62.4 & 67.6 \\
GPT-4o                       & 80.9 & 58.5 & \textbf{67.9}  & 88.0 & 67.8 & 76.6  & 70.8 & 62.1 & 66.2  & 72.8 & 59.2 & 65.3 & 69.0 \\
GPT-4o-mini                  & 62.1 & 54.1 & 57.8  & 68.3 & 61.6 & 64.8  & 52.1 & 60.4 & 55.9  & 39.4 & 56.1 & 46.3 & 56.2 \\
o1-mini                      & 82.6 & 54.6 & 65.7  & 84.7 & \textbf{74.3} & \textbf{79.2}  & 77.2 & \textbf{71.3} & \textbf{74.2}  & \textbf{84.5} & \textbf{72.9} & \textbf{78.3} & \textbf{74.4} \\
\midrule
\textbf{Average} & 70.7 & 53.9 & 58.7  & 73.5 & 63.5 & 66.3  & 65.4 & 61.5 & 61.2  & 60.2 & 59.9 & 57.1 & 60.8 \\
\bottomrule
\end{tabular}
}
\caption{Detailed performance of PRMs and LLM-as-a-judge models across the four subsets of \textsc{ReTraceQA}. Each triplet reports accuracy on identifying correct reasoning traces (correct), accuracy on pinpointing the exact error location (error), and overall F1 score. The final column reports the average F1 score across all subsets.}
\label{tab:retrace-results}
\end{table*}

\subsection{Reasoning Trace Evaluation}\label{sec:res-trace-eval}

\paragraph{Reference-free Evaluation.}\label{sec:reference-free}
Table \ref{tab:retrace-results} (top) highlights substantial limitations of LLM-as-a-judge and PRMs when asked to assess SLM reasoning traces in commonsense reasoning tasks under reference-free evaluation.
Overall, F1 scores across all four datasets remain relatively low, even for the strongest judges, suggesting that reliably evaluating the soundness of reasoning traces remains a challenging task. 
While state-of-the-art LLMs like GPT-4o and o1-mini outperform others with F1 scores exceeding 60\% on some datasets, the average F1 across models hovers around 54–56\%, indicating considerable room for improvement. 
These results underscore a key challenge in deploying LLM-as-a-judge models in reinforcement learning or Best-of-N selection settings: their current inability to robustly identify and reward correct intermediate reasoning without having access to the correct label limits their usefulness for guiding reasoning-focused learning objectives.

Additionally, PRMs originally developed and trained for mathematical tasks perform significantly worse across all datasets, with average F1 scores often below 25\%. 
This performance gap emphasizes the PRMs' limited generalization capabilities when transferred to commonsense reasoning (see \citet{pisano-etal-2026-process} for extending the logical capabilities of PRMs).

\begin{figure}[t]
    \centering
    \begin{subfigure}{\linewidth}
        \centering
        \includegraphics[width=\linewidth]{benchmark_error_steps.png}
        \caption{Human-annotated step error positions.}
        \label{fig:benchmark-error-steps}
    \end{subfigure}

    \vspace{1em} 

    \begin{subfigure}{\linewidth}
        \centering
        \includegraphics[width=\linewidth]{judge_o1-mini_error_steps.png}
        \caption{o1-mini step error positions.}
        \label{fig:judge-error-steps}
    \end{subfigure}

    \caption{Comparison of step error positions: (a) human annotation and (b) o1-mini employed as a judge.}
    \label{fig:combined-error-steps}
\end{figure}

\begin{table}[t]
\centering
\resizebox{0.48\textwidth}{!}{
\begin{tabular}{lcccc}
\toprule
\textbf{Dataset} & \multicolumn{2}{c}{\textbf{o1-mini (ext.)}} & \multicolumn{2}{c}{\textbf{o1-mini (judge)}} \\
& accuracy & err. rec. & accuracy & err. rec. \\
\midrule
CSQA          & 82.2 & 65.7 & 81.9 & 81.1 \\
OBQA          & 84.8 & 74.3 & 90.2 & 94.3 \\
QASC          & 83.0 & 74.1 & 86.2 & 91.5 \\
StrategyQA    & 74.8 & 62.8 & 90.0 & 92.1 \\
\midrule
Average       & 81.2 & 69.2 & \textbf{87.0} & \textbf{89.8} \\
\bottomrule
\end{tabular}
}
\caption{Accuracy and error recall (\%) of o1-mini employed as answer extractor (ext.) and as judge on \textsc{ReTraceQA}. Accuracy measures correct trace classification; error recall measures erroneous trace detection.}
\label{tab:judge_accuracy}
\end{table}

\begin{table*}[t]
\centering
\resizebox{\textwidth}{!}{
\begin{tabular}{l|ccc|ccc|ccc|ccc|ccc}
\toprule
\textbf{Model} &
\multicolumn{3}{c|}{\textbf{CommonsenseQA}} &
\multicolumn{3}{c|}{\textbf{OpenBookQA}} &
\multicolumn{3}{c|}{\textbf{QASC}} &
\multicolumn{3}{c|}{\textbf{StrategyQA}} &
\multicolumn{3}{c}{\textbf{Avg. Accuracy}} \\
\textit{(Instruct)} & ext. & judge & $\Delta$ & ext. & judge & $\Delta$ & ext. & judge & $\Delta$ & ext. & judge & $\Delta$ & ext. & judge & $\Delta$ \\
\midrule
Llama-3.2-1B     & ~47.6 & ~27.7 & ~19.9 & ~47.2 & ~23.8 & ~23.4 & ~47.7 & ~22.2 & ~25.5 & ~53.7 & ~20.1 & ~33.6 & ~49.0 & ~23.4 & ~25.6 \\
Llama-3.2-3B     & ~69.1 & ~58.7 & ~10.4 & ~75.4 & ~58.4 & ~17.0 & ~71.9 & ~51.4 & ~20.5 & ~64.3 & ~34.8 & ~29.5 & ~70.2 & ~50.8 & ~19.4 \\
Llama-3.1-8B     & ~75.7 & ~72.8 & ~~~2.9 & ~83.0 & ~69.6 & ~13.4 & ~79.4 & ~64.0 & ~15.4 & ~67.2 & ~45.9 & ~21.3 & ~76.3 & ~63.1 & ~13.2 \\
Phi-4-Mini      & ~65.8 & ~57.1 & ~~~8.7 & ~79.8 & ~65.0 & ~14.8 & ~67.2 & ~49.1 & ~18.1 & ~60.1 & ~42.4 & ~17.7 & ~68.2 & ~53.4 & ~14.8 \\
Qwen2.5-1.5B     & ~67.1 & ~45.9 & ~21.2 & ~62.6 & ~40.8 & ~21.8 & ~57.0 & ~33.2 & ~23.8 & ~57.0 & ~27.8 & ~29.2 & ~60.9 & ~36.9 & ~24.0 \\
Qwen2.5-3B       & ~74.2 & ~60.8 & ~13.4 & ~77.6 & ~53.8 & ~23.8 & ~71.9 & ~48.2 & ~23.7 & ~58.1 & ~31.0 & ~27.1 & ~70.4 & ~48.5 & ~22.0 \\
Qwen2.5-7B      & ~82.4 & ~78.5 & ~~~3.9 & ~87.8 & ~72.2 & ~15.6 & ~81.4 & ~67.8 & ~13.6 & ~72.5 & ~51.4 & ~21.1 & ~81.0 & ~67.5 & ~13.5 \\
\midrule
\textbf{Average}                  & ~68.8 & ~57.4 & ~11.4 & ~73.3 & ~54.8 & ~18.5 & ~67.9 & ~48.0 & ~19.9 & ~61.9 & ~36.2 & ~25.7 & ~68.3 & ~49.7 & ~18.6 \\
\bottomrule
\end{tabular}
}
\caption{Accuracy (\%) of seven SLMs on four commonsense benchmarks, evaluated using o1-mini as an answer extractor (ext.) and as a judge. $\Delta$ represents the performance inflation introduced by answer-only evaluation.}
\label{tab:slm_final_accuracy}
\end{table*}

\paragraph{Reference-based Evaluation.}\label{sec:reference-based}
Table \ref{tab:retrace-results} (bottom) reports reference-based evaluation results across the four subsets in \textsc{ReTraceQA}. 
Most models achieve moderate to strong performance in identifying globally correct reasoning traces, but accurately localizing specific error steps remains substantially more challenging. 
Model size correlates positively with performance. For instance, Qwen2.5-72B-Instruct outperforms Mistral-Small-24B-Instruct by +35.5\% F1 on average. 
However, scale alone is insufficient: DeepSeek-R1, despite being larger than Qwen2.5-72B-Instruct, underperforms across all datasets, suggesting that architectural choices and reasoning-oriented training are critical. 
The strongest judge, o1-mini, achieves 74.4\% F1, highlighting the importance of effective reasoning-oriented objectives.
Moreover, we can see that a consistent pattern emerges: models detect trace correctness better than localizing errors. 
For instance, o1-mini achieves 74.3\% error classification on OBQA, still lagging behind its correctness detection ability. 
Figure \ref{fig:combined-error-steps} compares error position distributions between human annotations and o1-mini predictions. 
Errors most commonly occur at steps 3-4, suggesting that, while early context establishment succeeds, errors emerge during mid-level inference. 
We can see that o1-mini's predictions mirror human patterns well, particularly on CSQA and QASC. But they show heavier tails, indicating over-assignment of blame to later steps, potentially capturing error consequences rather than origins.
These results highlight both the promise and limitations of LLM-as-a-judge systems: while stronger models align well with human evaluations of overall correctness, precisely identifying error origins remains an open challenge.

\subsection{Downstream SLM Evaluation}\label{sec:slm_eval}

While the reference-based evaluation results in Section \ref{sec:res-trace-eval} show that current LLMs employed as automated judges may not reliably localize errors for fine-tuning feedback, their strong performance in assessing overall trace correctness suggests potential for more reliable SLM evaluation.
Here we investigate whether reasoning-aware judges that consider both trace validity and final answers provide more accurate assessments than standard answer-only evaluation.

Table \ref{tab:judge_accuracy} demonstrates that the best-performing judge on \textsc{ReTraceQA} (o1-mini) employed as a reasoning-aware judge consistently outperforms standard answer extraction, achieving +5.8 points in accuracy (correctly distinguishing valid from invalid traces) and +20.6 points in error recall (identifying flawed traces) on average.
This emphasizes the importance of process-aware evaluation for faithful SLM assessment.

Building on this finding, we evaluate multiple SLMs under both paradigms to quantify the discrepancy between answer-only and reasoning-aware evaluation. 
Table \ref{tab:slm_final_accuracy} reveals a consistent 18.6 percentage point average drop when using reasoning-aware evaluation, demonstrating how traditional metrics overestimate SLM capabilities. 
Even high-performing models like Qwen2.5-7B-Instruct show substantial drops (81.0\% to 67.5\%). 

These results align with our benchmark analysis (Section \ref{sec:benchmark-stats}) showing that 17.9\% of instances reach correct answers through flawed reasoning, and reinforce that o1-mini as a judge better aligns with human assessments. 
These findings underscore the critical need for reasoning-aware evaluation frameworks that move beyond final answer correctness to accurately reflect SLM reasoning capabilities.

\section{Conclusions}
In this work, we introduced \textsc{ReTraceQA}, a new gold benchmark for evaluating reasoning traces of SLMs through step-level annotations, including error category locations and categorizations.
Our manually annotated benchmark reveals that standard answer-only metrics consistently overestimate SLM performance: on average, 17.9\% of the time, SLMs arrive at correct answers via reasoning that contains at least one significant error. Moreover, introducing reasoning-aware evaluation shows that their scores are inflated by up to 25\%.
Our manual error analysis shows that SLMs struggle primarily with factual grounding (hallucinations account for 41.9-62.5\% of errors), though logical coherence issues are also significant (27.9-35.4\%).

Although our work demonstrates that LLMs are strong judges and can distinguish correct vs. incorrect traces effectively, they still struggle with error localization. 
Additionally, PRMs trained on math reasoning fail to transfer to commonsense tasks, highlighting domain-specific gaps and the need for non-math process-level benchmarks like \textsc{ReTraceQA}.
We hope that \textsc{ReTraceQA} will encourage broader adoption of reasoning-aware evaluation protocols, thereby providing more reliable assessments of language models.

\section*{Limitations}
Our work provides valuable insights into reasoning trace evaluation for commonsense reasoning, though some limitations should be taken into account.
First, our benchmark focuses exclusively on English-language commonsense reasoning tasks.
Extending this evaluation framework to multilingual settings would be valuable for understanding whether reasoning patterns and error distributions vary across languages.
Second, while we selected four diverse datasets for commonsense reasoning and extended reasoning trace evaluation beyond mathematics and science, it would be valuable to extend current work on benchmarks capturing reasoning patterns required in other domains, such as procedural reasoning or narrative comprehension. 
Future work should extend this evaluation framework to a broader range of reasoning modalities in order to establish more comprehensive benchmarks.
Finally, our results demonstrate that PRMs trained on mathematical reasoning transfer poorly to commonsense domains, with performance degrading substantially.
This domain transfer limitation suggests that reasoning evaluation techniques may require domain-specific adaptations rather than assuming general transferability across reasoning tasks, motivating the need to develop specialized PRMs for other domains beyond mathematics and science.

Overall, while our work provides a solid foundation for reasoning trace evaluation, addressing the aforementioned limitations will be crucial for advancing the field and developing more robust reasoning models.

\section*{Acknowledgments}
\begin{center}
\noindent
    \begin{minipage}{0.1\linewidth}
        \begin{center}
            \includegraphics[scale=0.05]{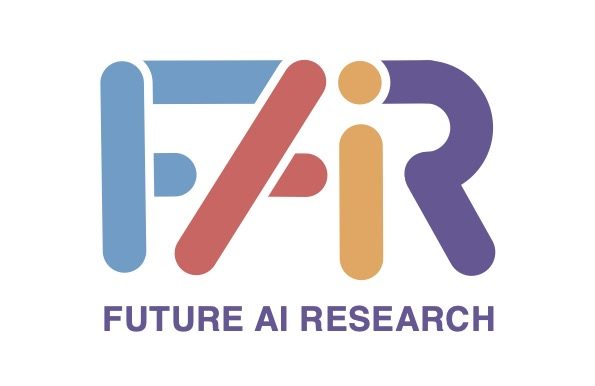}
        \end{center}
    \end{minipage}
    \hspace{0.01\linewidth}
    \begin{minipage}{0.70\linewidth}
        Roberto Navigli and Simone Conia gratefully acknowledge the support of the PNRR MUR project PE0000013-FAIR. Simone's fellowship is fully funded by this project.
    \end{minipage}
    \hspace{0.01\linewidth}
    \begin{minipage}{0.1\linewidth}
        \begin{center}
            \includegraphics[scale=0.08]{eu.png}
        \end{center}
    \end{minipage}\\
\end{center}
\vspace{0.2cm}

\bibliography{ms}

\appendix

\section{Solution Generation Prompts}\label{app:cot-prompts}
Table \ref{tab:prompt-zs-cot-mcqa} and \ref{tab:prompt-zs-cot-binary} show the prompts used in our work to generate the solutions for the multiple-choice and binary commonsense reasoning benchmarks, respectively. 
Specifically, we use the standard zero-shot Chain-of-Thought (CoT) prompting strategy \cite{wei2023chainofthoughtpromptingelicitsreasoning, kojima2023largelanguagemodelszeroshot} to elicit explicit model reasoning.

\section{Answer Extractor Details}\label{app:answer-extractor}
To ensure that our benchmark contains a balanced number of model outputs reaching a correct or incorrect solution, we use a state-of-the-art LLM-based answer extractor \cite[xFinder]{yu2025xfinderlargelanguagemodels}.
We select this answer-extraction method rather that relying on simple regular expressions because of its higher agreement with human judgment in scenarios involving free-form text generation \cite{molfese-etal-2025-right}. 
Specifically, we adopt xFinder-llama38it,\footnote{\url{https://huggingface.co/IAAR-Shanghai/xFinder-llama38it}} the best-performing variant based on Meta-Llama-3-8B-Instruct.
We prompt xFinder by providing it with the input question, the optional set of choices and the model output.
We then extract its generated output and compare it against the correct answer.
We deem an instance as correct if it reaches the correct answer and incorrect otherwise.

\section{Error Examples}\label{app:error-examples}
Table \ref{tab:additional-reasoning-trace-errors} shows additional examples drawn from our \textsc{ReTraceQA} benchmark.
Specifically, the table presents the input problem (consisting of a question and an optional set of choices), the model’s reasoning trace divided into paragraphs, and the category of error, annotated according to one of three categories.
This table provides a qualitative overview of the errors made by SLMs in the context of commonsense reasoning.
In particular, the table highlights problematic reasoning patterns in the models’ outputs, with errors annotated following the categorization defined in Section \ref{sec:human-annotation}.
The first example represents a misinterpretation error, where the model fails to relate the meaning of the answer choices to the question.
The second example shows an hallucination error, where the reasoning traces contains the incorrect statement ``bowling alley is not the typical location for throwing a ball at pins''.
Finally, the third example illustrates a reasoning error in which the model correctly states ``A. ecru: This color is very light, almost white color'' but then continues with an illogical consequence which is ``It would not provide much heat-reflecting capability''.

\begin{table}[t]
    \renewcommand{\arraystretch}{1.6}
    \centering
    \begin{tabular}{|p{0.94\linewidth}|}
        \hline
        \textbf{SYSTEM} 
        \\
        \texttt{You are an expert in commonsense question answering. You are given as input a question and a set of choices. First, provide the reasoning process to answer the question. Finally, provide your final answer.}
        \\
        \hline
        \textbf{USER} 
        \\
        \texttt{Question: \{question\}}
        \\[-8pt]
        \texttt{Choices: \{choices\}}
        \\
        \hline
    \end{tabular}
    \caption{Prompt for CSQA, OBQA and QASC datasets.}
    \label{tab:prompt-zs-cot-mcqa}
\end{table}

\begin{table}[t]
    \renewcommand{\arraystretch}{1.6}
    \centering
    \begin{tabular}{|p{0.94\linewidth}|}
        \hline
        \textbf{SYSTEM} 
        \\
        \texttt{You are an expert in commonsense question answering. You are given as input a yes/no question. First, provide the reasoning process to answer the question. Finally, provide your final answer as 'yes' or 'no'.}
        \\
        \hline
        \textbf{USER} 
        \\
        \texttt{Question: \{question\}}
        \\
        \hline
    \end{tabular}
    \caption{Prompt for the StrategyQA dataset.}
    \label{tab:prompt-zs-cot-binary}
\end{table}

\section{Annotation Guidelines}\label{app:annotation-guidelines}

\subsection{Task Overview}

The goal of our annotation process is to evaluate the step-by-step reasoning traces generated by SLMs in response to questions requiring commonsense reasoning. 
Annotators are tasked to identify the earliest point in the reasoning trace where an error occurs. 
Importantly, the final answer produced by the model may be correct even if the intermediate reasoning steps are flawed. 
As such, annotations focus solely on the reasoning trace rather than final output alone.
Each annotation instance includes the question, optional answer choices (for multiple-choice tasks), the ground truth answer, and the model-generated reasoning trace broken into discrete steps.\footnote{In the following, we use the terms ``steps'' and ``paragraphs'' interchangeably.} In some cases, additional contextual facts drawn from the original datasets are provided to assist the annotator.

\subsection{Annotation Objective}

Annotators are instructed to read each step in the reasoning trace and identify the first step that contains an error.
The task is framed as a classification problem, where annotators assign an integer index to indicate the position of the first erroneous step. 
Step indices are zero-based (i.e., $0$ refers to the first step), and a value of $-1$ is used to denote that all steps in the reasoning trace are correct.
To minimize the inherent subjectivity of commonsense judgment and provide a reproducible framework for evaluation, annotators apply a hierarchical decision tree. This ensures that every reasoning failure is categorized using one of three mutually exclusive labels based on the structural level of the error: misinterpretation, hallucination and reasoning.

\subsection{Error Definition}

We define three categories of errors based on the structural stage at which the reasoning process fails. To remove ambiguity, annotators are instructed to evaluate these categories hierarchically, ensuring that the primary cause of the failure is identified at the most fundamental level (from grounding to inference).

\paragraph{Misinterpretation (Grounding Level).}
This is the primary level of the taxonomy, representing errors where the model fails to correctly ground its reasoning in the provided input or task constraints. By identifying these errors first, we isolate failures in task comprehension from failures in knowledge or logic.
This includes:
\begin{itemize}
    \item Misinterpreting the core question objective or task requirements.
    \item Misrepresenting previous reasoning steps or established premises.
    \item Referencing non-existent answer choices or providing multiple answers when only one is requested.
\end{itemize}

\textit{Example:} For the question ``Why would you take a bus to work?'' with choice A being ``commute,'' a model ruling out this correct option because ``the question asks why someone would take a bus, not what a bus is used for'' demonstrates misinterpretation of the question's objective.

\paragraph{Hallucination (Content Level).}
If the input is correctly grounded, we evaluate the accuracy of the external knowledge introduced. This category is strictly reserved for the introduction of empirically false or unverifiable world knowledge. It focuses on the \textit{factual accuracy} of the information used in the trace, independent of the logical structure.
This includes:
\begin{itemize}
    \item Introduction of incorrect facts or assumptions that are not generally valid.
    \item Generation of hallucinated information that is not inferable from the question or context.
\end{itemize}

\textit{Example:} A model stating ``rejection is the most likely outcome of an interview'' presents an incorrect fact that is not generally valid, constituting a hallucination error.

\paragraph{Reasoning (Inference Level).}
This category addresses errors in the logical transitions between or within steps. It is applied only when the grounding and factual content are correct, but the model connects them using an invalid logical leap or contradictory inference. This ensures that logical failures are evaluated as pure errors in deduction.
This includes:
\begin{itemize}
    \item Logically unsound or commonsense-violating inferences.
    \item Contradictory or internally incoherent reasoning.
    \item Ruling out the correct option despite valid intermediate steps.
\end{itemize}

\textit{Example:} For the question ``Where spiders might be found among tools?'', a model stating ``a garage may store tools'' but then ruling out garage as ``not the most likely place'' with unsound reasoning commits a reasoning error.

\subsection{Non-errors}

Not all irregularities in reasoning traces qualify as errors. Annotators are explicitly instructed \textit{not} to flag the following as erroneous:

\begin{itemize}
    \item Minor grammatical issues or unusual phrasing that do not affect semantic content.
    \item Verbose, redundant, or overly detailed reasoning that remains logically sound.
\end{itemize}

\subsection{Annotation Procedure}

The annotation process involves six key steps:

\begin{enumerate}
    \item Read the question, any associated answer choices, and any additional supporting facts.
    \item Consult the ground truth answer to understand the correct resolution.
    \item Examine each reasoning step in sequence.
    \item Determine whether each step is sound following the provided guidelines.
    \item Record the index of the first erroneous step, or $-1$ if all steps are correct.
    \item Categorize the error (hallucination, reasoning, or misinterpretation) if an error is found.
\end{enumerate}

Importantly, annotators are instructed to mark only the earliest point in the trace where an error occurs, as later steps may be incorrect solely due to propagation from a previously erroneous step \cite{lightman2023letsverifystepstep, zheng2024processbenchidentifyingprocesserrors}.
This process ensures that annotations are consistent, fine-grained, and focused on evaluating the internal validity of reasoning traces rather than their final outcomes alone.

\subsection{Identifying Invalid Instances}

During the annotation process, some instances may be identified as problematic due to flaws in the original question or labeling. 
To maintain the quality of the benchmark and avoid propagating errors from upstream datasets, annotators are asked to flag such examples with a special \textsc{Invalid} tag. 
The following conditions qualify an instance for exclusion: (i) the question contains severe grammatical issues that compromise its interpretation; (ii) multiple answer choices are semantically identical or equally valid; (iii) the dataset's annotated ground truth is demonstrably incorrect based on commonsense or factual knowledge; and (iv) the instance lacks a unique, clearly correct answer.
Instances tagged as \textsc{Invalid} are excluded from the analysis.


\section{Individual Model Statistics}\label{app:model-stats}

Figure \ref{fig:individual-model-error_rate} shows the individual process error rates of the seven SLMs listed in Section \ref{sec:solution-generation} (we omit ``Instruct'' tags for readability), measured as the percentage of instances in which an incorrect reasoning trace leads to the correct final answer.
In particular, we can see that the trend is approximately the same for each subset of \textsc{ReTraceQA}: smaller models correspond to higher process error rates, with the latter decreasing as model size increases.

\section{LLM-as-a-judge Prompts}\label{app:judge-prompts}
Table \ref{tab:judge-prompt} show the prompts used for LLM-as-a-judge evaluation of commonsense reasoning traces. 
They are lightly adapted from prior work~\cite{zheng2024processbenchidentifyingprocesserrors}, omitting answer choices when unavailable and including supporting facts when provided.

\begin{table*}[t]
    \begin{adjustbox}{width=\textwidth}
        \footnotesize
        \centering
        \begin{tabular*}{\textwidth}{p{0.25\textwidth}|p{0.50\textwidth}|p{0.17\textwidth}}
            \toprule
            \textbf{Input} & \textbf{Reasoning Trace} & \textbf{Error Category} \\ 
            \midrule

            \parbox[t]{0.25\textwidth}{
                \textbf{Question}: birds use their peckers to catch \\
                \textbf{Choices}: \\
                A. dogs \\
                B. a tan \\
                C. a ball \\
                \underline{D. bees} \\
            } &
            \parbox[t]{0.50\textwidth}{                
                \textbf{Paragraph 0:} \par
                The question asks what birds use their peckers for. \par
                [...] \par
                \textbf{Paragraph 5:} \par
                \parbox{0.49\textwidth}{
                    \textcolor{red}{The choices provided are not related to the question at hand.}
                }
            } &
            \parbox[t]{0.17\textwidth}{Misinterpretation.} \\

            \midrule
            
            \parbox[t]{0.25\textwidth}{
                \textbf{Question}: Where do you throw a ball at pins? \\
                \textbf{Choices}: \\
                \underline{A. bowling alley} \\
                B. football stadium \\
                C. soccer field \\
                D. sporting event \\
                E. sporting goods store
            } &
            \parbox[t]{0.50\textwidth}{
                \textbf{Paragraph 0:} \par
                To answer this question, we need to consider the context and the typical location of a ball-throwing activity. \par
                [...] \par
                \textbf{Paragraph 2:} \par
                \parbox{0.49\textwidth}{
                    \textcolor{red}{A bowling alley is primarily used for bowling, a specific type of ball-throwing game. It is not the typical location for throwing a ball at pins.}
                }
            } &
            \parbox[t]{0.17\textwidth}{Hallucination.} \\
            
            \midrule

            \parbox[t]{0.25\textwidth}{
                \textbf{Question}: What is a more comfortable color to have for your automobile upholstery if living in a desert? \\
                \textbf{Choices}: \\
                \underline{A. ecru} \\
                B. red \\
                C. black \\
                D. navy
            } &
            \parbox[t]{0.50\textwidth}{             
                \textbf{Paragraph 0:} \par
                To answer this question, we need to consider the climate and environment of a desert. Deserts typically have high temperatures and intense sunlight, so the upholstery color should ideally have good heat-reflecting properties to keep the car interior cooler. \par
                [...] \par
                \textbf{Paragraph 2:} \par
                \parbox{0.49\textwidth}{
                    \textcolor{red}{A. ecru: This color is a very light, almost white color. It would not provide much heat-reflecting capability.}
                }
            } & 
            \parbox[t]{0.17\textwidth}{Reasoning.} \\

            \bottomrule
        \end{tabular*}
    \end{adjustbox}
    \caption{Additional examples drawn from our \textsc{ReTraceQA} benchmark.}
    \label{tab:additional-reasoning-trace-errors}
\end{table*}

\begin{figure*}[t]
 \begin{subfigure}{0.49\textwidth}
     \includegraphics[width=\textwidth]{process_error_rate_commonsense_qa.png}
     \caption{CSQA Process Error Rate}
     \label{fig:error_rate_commonsesnseqa}
 \end{subfigure}
 \hfill
 \begin{subfigure}{0.49\textwidth}
     \includegraphics[width=\textwidth]{process_error_rate_openbookqa.png}
     \caption{OBQA Process Error Rate}
     \label{fig:error_rate_openbookqa}
 \end{subfigure}
 
 \par\medskip
 
 \begin{subfigure}{0.49\textwidth}
     \includegraphics[width=\textwidth]{process_error_rate_qasc.png}
     \caption{QASC Process Error Rate}
     \label{fig:error_rate_qasc}
 \end{subfigure}
 \hfill
 \begin{subfigure}{0.49\textwidth}
     \includegraphics[width=\textwidth]{process_error_rate_StrategyQA.png}
     \caption{StrategyQA Process Error Rate}
     \label{fig:error_rate_strategyqa}
 \end{subfigure}
  \caption{Process Error Rate (\%): The proportion of incorrect reasoning traces that reach the correct final answer, calculated across the annotated subsets of our benchmark for each model.}
 \label{fig:individual-model-error_rate}
\end{figure*}

\begin{table*}[t]
    \renewcommand{\arraystretch}{1.6}
    \centering
    \begin{tabular}{|p{0.94\linewidth}|}
        \hline
        \textbf{SYSTEM} \\
        You are an expert in carefully analyzing step-by-step solutions for commonsense reasoning problems. \\
        \hline
        \textbf{USER} \\
        The following is a commonsense reasoning problem composed of a question, a set of choices, the correct answer and a solution (split into paragraphs, enclosed with tags and indexed from 0): \\
        \textbf{[Commonsense Problem]} \\
        Question: \{\texttt{question}\} \\
        Choices: \{\texttt{choices}\} \\
        Answer: \{\texttt{answer}\} \\
        \textbf{[Solution]} \\
        \{\texttt{model\_output}\} \\
        Your task is to review and critique the solution paragraph by paragraph. Once you identify a commonsense reasoning error in a paragraph, return the index of the paragraph where the earliest error occurs. Otherwise, return the index of -1 (which typically denotes `not found`). \\
        Please put your final answer (i.e., the index) in \texttt{boxed\{\}}. \\
        \hline
    \end{tabular}
    \caption{LLM-as-a-judge prompt.}
    \label{tab:judge-prompt}
\end{table*}

\end{document}